\pdfoutput=1

\documentclass[11pt]{article}

\usepackage{acl}
\usepackage{times}
\usepackage{latexsym}
\usepackage{float} 

\usepackage{calc}
\usepackage{titlesec}
\titlespacing*{\paragraph}{1pt}{1pt}{1em}

\usepackage[T1]{fontenc}
\usepackage[utf8]{inputenc}
\usepackage{microtype}
\usepackage{inconsolata}
\usepackage{graphicx}

\usepackage[font=small]{caption}

\usepackage{float} 

\usepackage{amsthm}

\usepackage{amsmath}
\usepackage[ruled,vlined]{algorithm2e}

\usepackage{enumitem}

\usepackage{multirow}
\usepackage{adjustbox}
\usepackage{bbm}
\usepackage{wrapfig,lipsum}
\usepackage{xspace}
\usepackage{booktabs,cellspace}  
\usepackage[normalem]{ulem}
\usepackage{makecell}
\usepackage{amssymb}
\usepackage{pifont}
\usepackage{todonotes}
\usepackage{microtype}

\usepackage{cleveref}

\usepackage{etoolbox}
\usepackage[tikz]{bclogo}

\usepackage{subcaption}
\usepackage{adjustbox}

\definecolor{msftBlue}{RGB}{0,164,239}
\definecolor{msftGreen}{RGB}{127,186,0}
\definecolor{msftYello}{RGB}{255,185,0}
\definecolor{msftBlack}{RGB}{0,0,0}

\usepackage{tcolorbox} 
\tcbuselibrary{skins} 
\usepackage[T1]{fontenc}

\tcbset{
    userstyle/.style={
        enhanced,
        colback=white,
        colframe=black,
        colbacktitle=gray!20,
        coltitle=black,
        rounded corners,
        sharp corners=north,
        boxrule=0.5pt,
        drop shadow=black!50!white,
        attach boxed title to top left={
            xshift=-2mm,
            yshift=-2mm
        },
        boxed title style={
            rounded corners,
            size=small,
            colback=gray!20
        }
    },
    replystyleg/.style={
        enhanced,
        colback=green!15,
        colframe=black,
        colbacktitle=green!30,
        coltitle=black,
        boxrule=0.5pt,
        drop shadow=black!50!white,
        rounded corners,
        sharp corners=north,
        attach boxed title to top right={
            xshift=-2mm,
            yshift=-2mm
        },
        boxed title style={
            rounded corners,
            size=small,
            colback=green!40
        }
    },
    replystyler/.style={
        enhanced,
        colback=red!15,
        colframe=black,
        colbacktitle=red!40,
        coltitle=black,
        boxrule=0.5pt,
        drop shadow=black!50!white,
        rounded corners,
        sharp corners=north,
        attach boxed title to top right={
            xshift=-2mm,
            yshift=-2mm
        },
        boxed title style={
            rounded corners,
            size=small,
            colback=red!40
        }
    }
}

\newtcolorbox{userquery}[1][]{
    userstyle,
    title=Prompt,
    #1
}

\title{Beyond Static Visual Tokens: Structured Sequential Visual Chain-of-Thought Reasoning}

\begin{document}
\maketitle
\begin{abstract}

Current multimodal LLMs encode images as static visual prefixes and rely on text-based reasoning, lacking goal-driven and adaptive visual access. Inspired by human visual perception—where attention is selectively and sequentially shifted from the most informative regions to secondary cues—we propose Structural Sequential Visual CoT \textbf{(SSV-CoT)}. First, a question-relevant saliency map identifies and organizes key visual regions, explicitly modeling the spatial distribution of visual importance. Second, reasoning is performed following this discriminative order, inducing a curriculum-like semantic progression from primary to secondary cues. This method is trained end-to-end, using text cot and answer supervision, without relying on region-level annotations or specialized external tools. Experiments on diverse visual reasoning benchmarks show gains, validating structured and sequential visual cognition. 


\end{abstract}

\section{Introduction}


Despite rapid advances in multimodal LLMs \citep{alayrac2022flamingovisuallanguagemodel, liu2023visualinstructiontuning, li2023blip2bootstrappinglanguageimagepretraining, dai2023instructblipgeneralpurposevisionlanguagemodels}, current systems still suffer from a fundamental limitation in how vision is integrated into reasoning. Most architectures encode an image once into visual tokens or a global embedding and inject it into the LLM as static context \citep{alayrac2022flamingovisuallanguagemodel, li2023blip2bootstrappinglanguageimagepretraining, liu2023visualinstructiontuning}. Reasoning then unfolds purely in text space: as the chain-of-thought grows, the model increasingly ignores the image, cannot revisit or operate on visual content, and lacks goal-directed visual actions such as selecting, refocusing, or comparing regions. As a result, visual perception and language reasoning remain largely disentangled, and vision functions as a passive prefix rather than an active part of cognition.

Existing attempts to mitigate this gap face significant drawbacks. External tool pipelines introduce cropping or detection but require complex orchestration \citep{wu2023visualchatgpttalkingdrawing, chen2023shikraunleashingmultimodalllms}. Region-based supervision depends on costly annotations \citep{lin2022reviveregionalvisualrepresentation, khan2022weaklysupervisedgroundingvqa}. Generative visual-token or image synthesis is inefficient and poorly aligned with reasoning needs \citep{esser2021tamingtransformershighresolutionimage, ding2022cogview2fasterbettertexttoimage}. Latent-space alignment methods learn global correspondences yet provide no explicit visual access steps \citep{radford2021learningtransferablevisualmodels, li2023blip2bootstrappinglanguageimagepretraining}. Interleaved chain-of-thought methods re-inject visual tokens but enable passive re-attention over a large patch space, without modeling where to look next \citep{zhang2024multimodalchainofthoughtreasoninglanguage}. None of these approaches support internal, sequential, goal-conditioned visual cognition.

\begin{figure}
    \centering
    \begin{adjustbox}{width=\linewidth}
        \includegraphics[width=\textwidth]{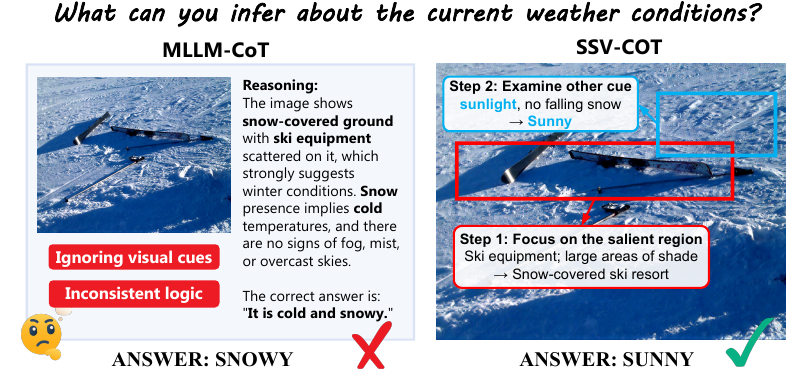}
    \end{adjustbox}
    \caption{MLLM lacks visual analysis, while SSV integrates cues through structured sequential reasoning to reach the correct answer.} 
    \label{fig:motivation}
\end{figure}

As illustrated in Fig.~\ref{fig:motivation}, humans rarely solve visual tasks by processing an entire scene uniformly. Instead, they first form a coarse global impression and then sequentially attend to task-relevant regions as reasoning unfolds, a pattern well documented in studies of human visual attention and eye movements \citep{yarbus1967eye,itti1998model}. In the example, the left approach (MLLM-CoT) fixates on the presence of snow-covered ground and ski equipment, ignores critical cues such as strong sunlight and sharp shadows, and arrives at an incorrect conclusion. In contrast, the right approach (SSV-CoT) explicitly guides visual attention through a structured sequence: it first focuses on salient regions (e.g., ski equipment and shadows) and then integrates additional cues (e.g., lighting conditions) to refine the hypothesis, ultimately reaching the correct answer. This comparison highlights a fundamental limitation of existing multimodal large language models, which typically compress images into a fixed set of visual tokens and perform reasoning purely in text space \citep{alayrac2022flamingovisuallanguagemodel,li2023blip2bootstrappinglanguageimagepretraining}. While recent work on chain-of-thought reasoning has shown the benefits of explicit intermediate reasoning in language models \citep{wei2023chainofthoughtpromptingelicitsreasoning, guo2025structuredoutputs}, visual perception in current MLLMs remains largely passive and decoupled from the reasoning process. This disconnect between perception and cognition motivates our central question: can a model learn not only \emph{what} to look at, but also \emph{when} and \emph{in which order} to look, driven directly by its ongoing reasoning?

Inspired by human visual cognition, we propose \textbf{SSV-CoT}, a multimodal reasoning framework that couples structured visual representation with sequential, goal-conditioned visual access. Given an image--question pair, SSV-CoT performs a single forward pass through a vision encoder and MLLM to construct a structured visual representation: a question-aware saliency map is generated, binarized, and decomposed via connected component analysis, yielding a bank of candidate region embeddings together with a global scene embedding. Conditioned on the evolving reasoning state, a visual cognition policy scores candidate regions and selects one region (or the global embedding) at each step, injecting its embedding into the MLLM, so that reasoning unfolds as an interleaved process of visual querying and textual inference. To reflect human attentional control, SSV-CoT further incorporates an adaptive stopping mechanism that dynamically determines whether to continue visual access, balancing evidence sufficiency against redundant queries.  

The method is trained end-to-end from image--question---text cot-answer without gaze traces, bounding boxes, or external tools: a heuristic supervised stage initializes the policy by imitating a question-conditioned saliency ordering derived from the pretrained MLLM, followed by a GRPO reinforcement learning stage that optimizes visual selection using answer-level rewards with a visual-budget regularizer. Overall, SSV-CoT replaces one-shot visual token injection with a structured, sequential, and adaptive visual reasoning process aligned with human cognitive principles.

Our main contributions are three-fold:
\vspace{-0.6em}
\begin{itemize}
  \setlength{\topsep}{0pt}
  \setlength{\itemsep}{0pt}
  \setlength{\parsep}{0pt}
  \setlength{\partopsep}{0pt}
  \item We model visual access in multimodal reasoning as a \emph{human-inspired, goal-conditioned} process over structured visual units, rather than passive visual token injection.
\item We propose a training framework that learns
visual selection order from task-level feed-
back using text-cot and answer supervision, without
region-level annotations.
\item We introduce an \emph{adaptive, budget-aware} visual reasoning mechanism that decides when to continue or stop visual querying during chain-of-thought reasoning.

\end{itemize}
\vspace{-0.6em}

\section{Related Work}

\subsection{Visual Reasoning}

A broad class of methods augments MLLMs with external visual tools for cropping, detection, or region selection \citep{schick2023toolformerlanguagemodelsteach, chen2023shikraunleashingmultimodalllms, yang2023mmreactpromptingchatgptmultimodal, aissi2025vipervisualperceptionexplainable,
guo-etal-2025-med}, but these rely on complex, non-differentiable pipelines with manually designed actions decoupled from internal reasoning. Another line of work introduces region-level supervision via bounding boxes, captions, or grounding annotations \citep{li2022groundedlanguageimagepretraining, yang2022unitabunifyingtextbox, kamath2021mdetrmodulateddetection, liu2024groundingdinomarryingdino}, which improves controllability but requires costly labeling and exposes all regions upfront, preventing sequential selection. Related segmentation-based or region-level abstractions, such as SAM-derived masks or region-level ViTs \citep{kirillov2023segment, chen2022regionvitregionaltolocalattentionvision, ravi2024sam2segmentimages}, provide structured features but are typically used as static inputs rather than a decision space. Overall, these approaches treat vision as fixed or externally controlled and do not enable internal, learnable, sequential visual cognition in MLLMs.

\subsection{Interleaved Visual Reasoning}

Interleaved chain-of-thought methods reintroduce visual tokens during reasoning to alleviate image forgetting \citep{zhang2024multimodalchainofthoughtreasoninglanguage, chen2025mintcot, shao2024visualcotadvancingmultimodal, zhang2024cocotcontrastivechainofthoughtprompting, li2025aimcotactiveinformationdrivenmultimodal}. While effective for multi-hop visual question answering and chart reasoning, these approaches repeatedly attend to the same static patch set and perform passive re-attention, without modeling where to look next or how visual access should evolve over time. Relatedly, sequential perception models such as glimpse networks and active-vision agents \citep{mnih2014recurrentmodelsvisualattention, ba2015multipleobjectrecognitionvisual, jaegle2022perceiveriogeneralarchitecture} demonstrate the benefits of goal-conditioned sequential inspection, but typically rely on small-scale vision architectures or reinforcement learning pipelines that do not scale to LLM-level multimodal reasoning. Overall, although recurrent and interleaved methods improve visual reasoning, they fail to support task-driven, autoregressive visual access that is tightly coupled with the reasoning process, leaving dynamic visual cognition largely unaddressed.

\section{Methodology}
\label{sec:method}

\begin{figure*}
    \centering
    \begin{adjustbox}{width=\linewidth}
\includegraphics[width=\textwidth]{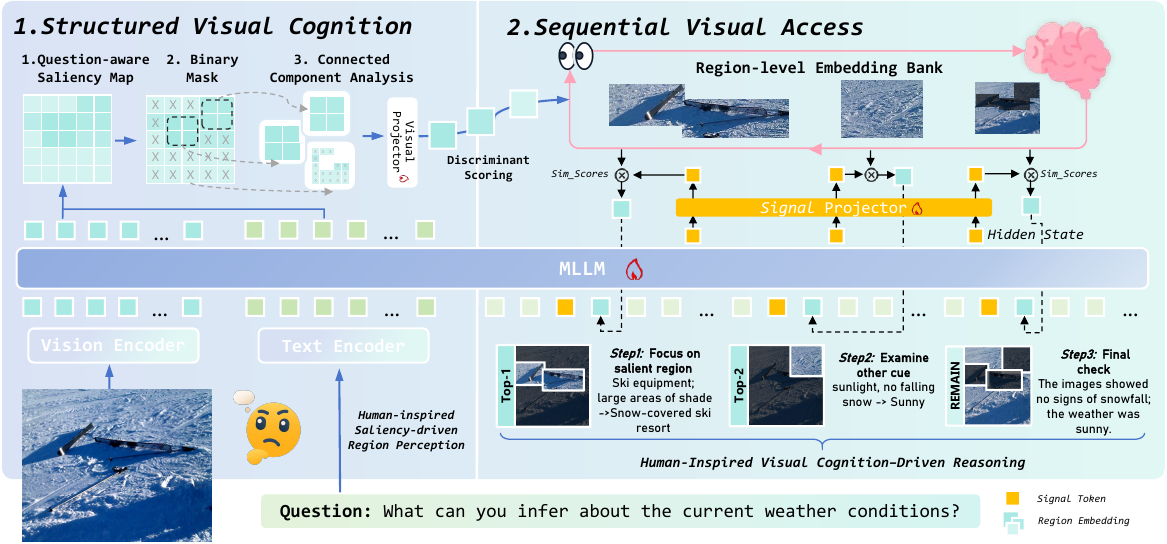}
    \end{adjustbox}
    \caption{Overview of the \textbf{SSV-CoT} framework. The model first constructs question-aware structured visual regions, then performs sequential visual access during chain-of-thought reasoning to progressively integrate visual evidence and generate the final answer. }
    \label{fig:vss}
\end{figure*}

\subsection{Overview}

Existing interleaved reasoning methods repeatedly inject dense patch-level visual tokens, often causing redundant signals and unstable reasoning. In contrast (Fig.~\ref{fig:vss}), we first abstract the image into a compact set of question-conditioned visual regions and then perform \emph{sequential visual access} over this region space. We introduce \textbf{SSV-CoT}, which decomposes multimodal reasoning into two stages: (1) \textbf{Structured Visual Cognition}, constructing salient question-aware region embeddings in a single forward pass; and (2) \textbf{Sequential Visual Access}, where a lightweight policy dynamically selects regions at each reasoning step.

\subsection{Problem Formulation and Inference Process}

Given an image $x$ and question $q_{\text{txt}}$, the model generates an answer by interleaving textual reasoning with selective visual access. Visual reasoning is formulated as a sequence of region-selection actions, where each step selects a region or terminates. At each reasoning step, the model queries a structured region embedding bank based on the current reasoning state until a \texttt{STOP} action produces the final answer.

Formally, while conventional multimodal CoT methods generate purely text-based reasoning steps
\begin{equation}
{s^{(1)}, s^{(2)}, \dots, s^{(K)}}, \quad y = \mathrm{LLM}(V, T),
\end{equation}
where $V={v_\tau}_{\tau=1}^N$ denotes static visual tokens extracted from the image and $T$ is the input question, our approach explicitly models reasoning as an interleaved decision process:
\begin{equation}
{a_1, s^{(1)}, a_2, s^{(2)}, \dots, a_T, s^{(T)}}, \quad y = \mathrm{LLM}(\mathcal{E}, T),
\end{equation}
where each action $a_t$ selects a region embedding from the structured visual space $\mathcal{E}$ (or triggers \texttt{STOP}), grounding each reasoning step in task-relevant visual evidence.

\subsection{Structured Visual Cognition}
\label{sec:proposal}

The first stage builds a compact, question-conditioned, and spatially structured image representation (Fig.~\ref{fig:vss}), aiming to identify a small set of discriminative regions relevant to the question without region-level supervision. Region geometry and base features come from the pretrained vision encoder, while region importance is inferred through question-conditioned multimodal interactions.

\paragraph{Question-conditioned saliency map.}
Given an image $x$, we extract patch-level features $h_v={z_{ij}}$ using a vision encoder, where $z_{ij}$ denotes patch embeddings on an $H\times W$ grid. We then run the pretrained MLLM on the image–question pair $(x, q_{\text{txt}})$ and obtain the final textual representation $q$ together with multimodal visual tokens $\tilde{z}_{ij}$. Since they lie in a shared embedding space, we compute a question-conditioned saliency score for each patch via
\begin{equation}
S_{ij}=\mathrm{sim}(q,\tilde{z}_{ij}),
\end{equation}
which reflects the relevance of each spatial location to the question. The saliency map is further normalized to improve robustness:
\begin{equation}
\tilde{S}{ij}=\frac{S_{ij}-\min(S)}{\max(S)-\min(S)}.
\end{equation}

\paragraph{Binary masking and connected component analysis.}
We convert the normalized saliency map $\tilde{S}$ into a binary mask by thresholding:
\begin{equation}
    M(i,j) = \mathbb{I}\big[\tilde{S}_{ij} \ge \tau\big],
\end{equation}
where $\tau$ is an adaptive threshold (e.g., Otsu’s method) and $\mathbb{I}[\cdot]$ denotes the indicator function.

We then apply connected-component analysis on $M$ under an 8-neighborhood connectivity criterion to obtain a set of candidate regions $\{\mathcal{R}_k\}$. Each region $\mathcal{R}_k$ consists of a connected set of salient patch indices. Small components below a minimum area threshold are discarded to remove spurious noise.

\paragraph{Global complement embedding.}
To ensure full image coverage and provide robustness against imperfect saliency estimation, all remaining patches not covered by the Top-$N$ regions are aggregated into a single global complement region:
\begin{equation}
    e_{\text{global}} = \frac{1}{|\mathcal{I}_{\text{rest}}|} \sum_{(i,j)\in \mathcal{I}_{\text{rest}}} z_{ij}.
\end{equation}

\paragraph{Discriminative region construction.}
Given the connected components $\{\mathcal{R}_k\}$ from the binary saliency mask, we treat each $\mathcal{R}_k$ as a candidate visual region and collect its patch-level visual tokens $\{v_j\}_{j=1}^{m}$. We adopt an adaptive token compression strategy with a fixed budget $n=48$. If $m \le n$, all tokens are preserved and projected individually via a visual projector. Otherwise, we select the top-$n$ tokens by saliency and apply $k$-means clustering to obtain representative region tokens. Formally, the region representation is a set of projected tokens:
\begin{equation}
E_k = g_{\mathrm{proj}}\big(\{z_{ij}\mid (i,j)\in \mathcal{R}_k\}\big), \quad k=1,\dots,N,
\end{equation}
where $g_{\mathrm{proj}}(\cdot)$ denotes a permutation-invariant region projector that produces a fixed-size set of region tokens.

After obtaining region embeddings, we quantify the discriminativeness of each
region by computing a region-level score as the average of the normalized
saliency values within the region:
\begin{equation}
    \rho_k = \frac{1}{|\mathcal{R}_k|} \sum_{(i,j)\in \mathcal{R}_k} \tilde{S}_{ij}.
\end{equation}
This score measures the overall saliency consistency of the region with respect
to the given question. Regions are then ranked according to $\rho_k$, and the Top-$N$ highest-scoring
regions are selected as discriminative local regions for subsequent sequential
visual reasoning. The final output of structured visual cognition is a compact region embedding bank:
\begin{equation}
    \mathcal{E} = \{e_1, \dots, e_N, e_{\text{global}}\},
\end{equation}
which provides a structured, question-conditioned abstraction of the image and serves as the candidate visual space for subsequent sequential visual access.

The structured region construction—including Otsu-based thresholding, binary masking, connected-component analysis, and Top-N selection—is treated as a deterministic, non-learnable preprocessing step performed at inference time, while all learnable parameters are confined to the projector, policy module, and the MLLM.

\subsection{Sequential Visual Access}
\label{sec:policy}

The second stage performs task-driven, step-aligned visual reasoning over the structured region space, corresponding to the right part of Figure~\ref{fig:vss}.

\paragraph{Step-aware triggering.}
During chain-of-thought decoding, we treat the newline character \texttt{\textbackslash n} as an implicit step boundary. Each time a newline is generated, the model completes a reasoning step and triggers a visual selection decision.

Let $h_t \in \mathbb{R}^{D_L}$ denote the final-layer hidden state of the newline token at step $t$. This hidden state summarizes the current reasoning context and serves as the query for visual selection.

\paragraph{Similarity-based region scoring.}
At each reasoning step $t$, we obtain the hidden state of the signal token $h_t \in \mathbb{R}^{D_L}$, which summarizes the current reasoning context. We first project the signal token through a signal projector:
\begin{equation}
\tilde{h}_t = W_{\mathrm{sig}}\, h_t \in \mathbb{R}^{D_L}.
\end{equation}
Similarly, each region token embedding $e_k \in \mathbb{R}^{D_v}$ is projected into the language space via a visual projector:
\begin{equation}
\tilde{e}_k = W_{\mathrm{vis}}\, e_k \in \mathbb{R}^{D_L}.
\end{equation}
We then compute similarity scores between the projected signal token and all projected region embeddings:
\begin{equation}
s_{t,k} = \mathrm{sim}(\tilde{h}_t, \tilde{e}_k), \quad k \in \{1,\dots,K\}.
\end{equation}
To enable adaptive termination of visual access, we additionally include a learnable \texttt{STOP} embedding, which is scored in the same manner as region tokens.

\paragraph{Policy and action execution.}
The policy head $g_\theta$ induces a distribution over actions:
\begin{equation}
    \pi_\theta(a_t = k \mid s_t) = \mathrm{softmax}(s_t)_k,
\end{equation}
where $k \in \{1,\dots,K,\text{STOP}\}$. If $a_t=k$, the corresponding region embedding is injected to condition subsequent reasoning; if $a_t=\text{STOP}$, the model produces the final answer without further visual input.

\subsection{Training Procedure}
\label{sec:training}

We train SSV-CoT with a two-stage protocol using image–question–textual CoT–answer data, without any supervision on visual regions or access order. To mitigate noisy early saliency estimates, we adopt a probabilistic curriculum that gradually shifts from random visual access to structured, discriminative region reasoning. The pipeline consists of (1) heuristic supervised fine-tuning to initialize the visual access policy aligned with textual chain-of-thought reasoning, and (2) GRPO-based reinforcement learning to refine the visual cognition order using answer-level rewards.

\paragraph{Stage I: Heuristic Supervised Fine-Tuning with Probabilistic Curriculum}
\label{sec:sft}

The goal of the first stage is to teach the model how to interleave textual
reasoning with visual region access, rather than to learn a perfect visual
selection strategy.
However, saliency maps estimated from a pretrained MLLM are often noisy at early
training stages, which may lead to unstable optimization if directly used as
supervision.

To stabilize training, we adopt a probabilistic curriculum learning strategy.
Specifically, we define a curriculum coefficient $\lambda_e \in [0,1]$ for epoch
$e$, which linearly increases from $0$ to $1$ during a warm-up period of
$E_{\text{warm}}$ epochs:
\begin{equation}
\lambda_e = \min\left(1, \frac{e}{E_{\text{warm}}}\right).
\end{equation}

For each training sample $(x, q_{\text{txt}}, a_{\text{gold}})$, we construct a
pseudo-expert region ordering as follows:
With probability $1 - \lambda_e$, we sample regions using a random
region strategy,
which selects spatially contiguous image blocks independent of the question.
With probability $\lambda_e$, we apply the proposed question-conditioned discriminative region selection method described in
Section~\ref{sec:proposal}, and rank the resulting regions by their saliency scores.

The resulting region sequence $(r_1, \dots, r_T)$ is treated as a heuristic
viewing trajectory.
During SFT, we perform behavior cloning on the visual access policy by minimizing
the cross-entropy loss:
\begin{equation}
\mathcal{L}_{\text{SFT}} =
- \sum_{t=1}^{T} \log \pi_\theta(a_t = r_t \mid s_t),
\end{equation}
where $s_t$ denotes the policy state at reasoning step $t$.

This curriculum-guided SFT initializes the policy with stable region--text
interleaving behavior while avoiding overfitting to unreliable early saliency
signals.

\paragraph{Stage II: GRPO-Based Reinforcement Learning with Mixed Exploration}

\label{sec:grpo}

While the heuristic policy learned in Stage I provides a stable initialization,
it is not directly optimized for end-task performance.
In the second stage, we refine the visual access policy using reinforcement
learning with answer-level supervision.

Given a training sample $(x, q_{\text{txt}}, a_{\text{gold}})$, we fix the image
and question and sample a group of $K$ trajectories using a mixed sampling
policy:
\begin{equation}
\pi_{\text{mix}} = (1 - \lambda_e)\,\pi_{\text{rand}} + \lambda_e\,\pi_\theta,
\end{equation}
where $\pi_{\text{rand}}$ denotes a random region selection policy and
$\pi_\theta$ is the current learned policy.
The same curriculum coefficient $\lambda_e$ is used to gradually shift from
exploration to exploitation.

Each trajectory $i$ yields a region access sequence $g^{(i)}$, a generated answer
$\hat{a}^{(i)}$, and a scalar reward $R^{(i)}$ computed solely based on answer
quality.
Within each group, we compute the group-relative advantage:
\begin{equation}
A^{(i)} = R^{(i)} - \frac{1}{K} \sum_{j=1}^{K} R^{(j)}.
\end{equation}

The policy is optimized using the GRPO objective:
\begin{equation}
\mathcal{L}_{\text{RL}} =
- \sum_{i=1}^{K} \sum_{t=1}^{T_i}
A^{(i)} \log \pi_\theta(a^{(i)}_t \mid s^{(i)}_t).
\end{equation}

To prevent large deviations from the SFT-initialized policy, we additionally
apply a KL regularization term:
\begin{equation}
\mathcal{L}_{\text{KL}} =
\mathbb{E}_{s}\!\left[
\mathrm{KL}\big(\pi_\theta(\cdot \mid s)\,\|\,\pi_{\text{SFT}}(\cdot \mid s)\big)
\right].
\end{equation}

The final optimization objective is:
\begin{equation}
\min_\theta \; \mathcal{L}_{\text{RL}} + \beta\,\mathcal{L}_{\text{KL}}.
\end{equation}

The reward is a linear combination of four terms:
\begin{equation}
r \;=\; r_{\text{task}} \;+\; \lambda_{1}\, r_{\text{format}} \;-\; \lambda_{2}\, r_{\text{length}} \;-\; \lambda_{3}\, r_{\text{vision}} .
\end{equation}
Here, $r_{\text{task}}$ is the task reward: for multiple-choice or numeric questions, it is $1$ if the answer is exactly correct and $0$ otherwise; for open-ended math reasoning, we use exact match with numerical-tolerance checking. The format reward $r_{\text{format}}$ encourages compliance with a predefined \emph{reasoning--answer} output template, and invalid formatting or missing a final answer is penalized. The length penalty $r_{\text{length}}$ is proportional to the number of generated tokens to discourage overly verbose reasoning. The vision penalty $r_{\text{vision}}$ is proportional to the number of visual access steps (or queried regions) to discourage excessive visual reliance and promote discriminative, budget-aware visual usage with adaptive stopping.

\section{Experiments}

\subsection{Experimental Datasets}

We evaluate our method on a diverse set of visual reasoning benchmarks.
For commonsense visual reasoning, we use M3CoT~\cite{zhang2024multimodalchainofthoughtreasoninglanguage}, ScienceQA~\cite{lu2022learnexplainmultimodalreasoning}, and LLaVA-Bench In-the-Wild (LLaVA-W)~\cite{liu2023visualinstructiontuning}.
M3CoT focuses on multi-domain, multi-step multimodal reasoning, ScienceQA serves as a standard benchmark for general VLM reasoning, and LLaVA-W evaluates fine-grained visual understanding via long-form answers with GPT-4V references.
For mathematical visual reasoning, we adopt MathVista~\cite{lu2024mathvistaevaluatingmathematicalreasoning} and MathVision~\cite{wang2024measuringmultimodalmathematicalreasoning}, which require both fine-grained visual perception and multi-step mathematical reasoning.

For training data, we use the M3CoT training set and the LLaVA-CoT-100k~\citep{xu2025llavacotletvisionlanguage} training set for commonsense visual reasoning. For mathematical visual reasoning, we adopt MathV360K~\citep{shi2024mathllavabootstrappingmathematicalreasoning}, from which we carefully select over 50K diverse samples and generate textual chain-of-thought annotations using GPT-4V. All the above datasets contain textual chain-of-thought annotations and do not include any visual grounding or visual token-level annotations.

\subsection{Compared Methods}
For \textbf{commonsense visual reasoning}, we compare our method with several representative reasoning strategies, all instantiated on the same \textsc{Qwen2-VL-7B} backbone for fairness. 
\textbf{Multimodal CoT}~\citep{zhang2024multimodalchainofthoughtreasoninglanguage} generates text-only intermediate reasoning steps before producing the final answer. 
\textbf{CCoT}~\citep{mitra2024compositionalchainofthoughtpromptinglarge} constructs a scene graph to capture compositional object relations and uses it to guide answer generation. 
\textbf{DDCoT}~\citep{zheng2023ddcotdutydistinctchainofthoughtprompting} decomposes the question into sub-questions and answers them using a VLM. 
\textbf{SCAFFOLD}~\citep{lei2024scaffoldingcoordinatespromotevisionlanguage} overlays coordinate grids onto images to expose fine-grained spatial information and guides reasoning with these coordinates. 
\textbf{ICoT}~\citep{gao2025interleavedmodalchainofthought} repeatedly re-injects visual tokens during CoT while treating visual patches uniformly.

For \textbf{mathematical visual reasoning}, we compare our method against a set of strong recent vision-language models, including MiniCPM-V-2.6~\cite{yao2024minicpmvgpt4vlevelmllm}, 
VITA-1.5~\cite{fu2025vita15gpt4olevelrealtime}, 
LLaVA-CoT~\cite{xu2025llavacotletvisionlanguage}, 
Qwen2-VL-7B~\cite{wang2024qwen2vlenhancingvisionlanguagemodels}, 
InternVL2.5~\cite{chen2025expandingperformanceboundariesopensource}, 
POINTS~\cite{liu2024points15buildingvisionlanguagemodel}, 
Ovis~\cite{lu2024ovisstructuralembeddingalignment}, 
and TVC-Qwen2-VL-7B~\cite{sun2025mitigatingvisualforgettingtakealong}. TVC-Qwen2-VL-7B mitigates visual forgetting in multimodal long-chain reasoning by periodically reintroducing compressed visual inputs during reasoning.

\subsection{Model Implementation }

We use Qwen2-VL-7B as the base model in our experiments. Both projectors are implemented as single-layer linear modules, and we uniformly set the similarity filtering threshold to $\theta = 0.7$. See the appendix for detailed settings.

\subsection{Overall Performance}

\begin{table}[t]
\centering
\small
\setlength{\tabcolsep}{4pt}
\begin{tabular}{lccc}
\toprule
\textbf{Methods} &
\textbf{M3CoT} &
\textbf{ScienceQA} &
\textbf{LLaVA-W} \\
& \textbf{ACC.↑} & \textbf{ACC.↑} & \textbf{ROUGE-L↑} \\
\midrule

Qwen2-VL-7B  & 43.6 & 56.3 & 32.7 \\
    ~~MultimodalCoT & 40.1 & 51.3 & 30.7 \\
    ~~CCoT & 43.3 & 56.4 & 29.4 \\
    ~~DDCoT & 42.6 & 55.2 & 31.2 \\
    ~~SCAFFOLD & 41.7 & 53.7 & 31.8 \\
    ~~ICoT & 44.1 & 56.8 & 34.2 \\
\midrule
\textbf{SSV-CoT (Ours)} & \textbf{44.9} & \textbf{57.3} & \textbf{35.7} \\
\bottomrule
\end{tabular}

\caption{
\textbf{Commonsense visual reasoning results.}
Comparison of SSV-CoT applied to Qwen2-VL-7B with CoT-based methods (including No-CoT, MultimodalCoT, CCoT, DDCoT, SCAFFOLD, and ICoT) on M3CoT (Accuracy), ScienceQA (Accuracy), and LLaVA-W (ROUGE-L). The best result for each benchmark is highlighted in bold.
}

\label{tab:qwen_vc_cot_0shot}
\end{table}

\begin{table}[t]
\centering
\footnotesize
\setlength{\tabcolsep}{2pt}
\begin{tabular}{lccc}
\toprule
\textbf{Model} & \textbf{MathVista} & \textbf{MathVision} & \textbf{Average} \\
               & \textbf{Acc.\,$\uparrow$} & \textbf{Acc.\,$\uparrow$} & \textbf{Acc.\,$\uparrow$} \\
\midrule

MiniCPM-V-2.6 & 60.8 & 18.4 & \textcolor{cyan!70!black}{39.6} \\
VITA-1.5 & 66.2 & 19.5 & \textcolor{cyan!70!black}{42.9} \\
LLaVA-CoT & 52.5 & 19.9 & \textcolor{cyan!70!black}{36.2} \\
Qwen2-VL-7B & 60.9 & 16.3 & \textcolor{cyan!70!black}{38.6} \\
InternVL2.5 & 64.5 & 17.0 & \textcolor{cyan!70!black}{40.8} \\
POINTS 1.5 & 66.4 & 22.0 & \textcolor{cyan!70!black}{44.2} \\
Ovis1.6-Gemma2 & 70.2 & 20.6 & \textcolor{cyan!70!black}{45.4} \\
TVC-Qwen2-VL-7B & 68.1 & 22.7 & \textcolor{cyan!70!black}{45.4} \\
\midrule
\textbf{SSV-CoT (Ours)} & \textbf{72.2} & \textbf{23.5} & \textbf{\textcolor{cyan!70!black}{47.9}} \\
\bottomrule
\end{tabular}

\caption{
\textbf{Mathematical visual reasoning results.}
Comparison of SSV-CoT applied to Qwen2-VL-7B with vision-language models on MathVista (Accuracy) and MathVision (Accuracy). 
The best results are highlighted in bold.
}
\label{tab:vc_cot_math}
\end{table}

Overall, the experimental results indicate that SSV-CoT leads to consistent improvements in long-context multimodal reasoning across both commonsense and mathematical tasks.

For commonsense visual reasoning (Table~\ref{tab:qwen_vc_cot_0shot}), SSV-CoT shows moderate gains over the Qwen2-VL-7B backbone, with improvements of +1.3 Accuracy on M3CoT, +1.0 Accuracy on ScienceQA, and +3.0 ROUGE-L on LLaVA-W. 
Compared with ICoT, which interleaves visual tokens while treating image patches uniformly, SSV-CoT better preserves task-relevant visual information through goal-conditioned and sequential visual selection, resulting in more stable performance across benchmarks.

For mathematical visual reasoning (Table~\ref{tab:vc_cot_math}), SSV-CoT also improves upon the Qwen2-VL-7B baseline, increasing the average score from 38.6 to 47.9. 
In comparison to TVC, which alleviates visual forgetting by periodically reintroducing compressed visual inputs, SSV-CoT further aligns visual selection with reasoning objectives, supporting more effective use of visual information in multi-step mathematical reasoning.

Overall, these results suggest that SSV-CoT helps mitigate visual forgetting and supports task-aware visual grounding, contributing to improved performance across diverse visual reasoning scenarios.

\section{Discussion}

\begin{table}[t]
\centering
\normalsize

\setlength{\tabcolsep}{4pt}
\begin{tabular}{lccc}
\toprule
\textbf{Methods} & M3CoT & ScienceQA & LLaVA-W \\
 & ACC.↑ & ACC.↑ & ROUGE-L↑ \\
\midrule
SSV-CoT & 44.9 & 57.3 & 35.7 \\
 ~~w/o SR & 43.5 (\textcolor{red}{-1.4}) & 56.1 (\textcolor{red}{-1.2}) & 33.9 (\textcolor{red}{-1.8}) \\
 ~~w/o SS & 43.8 (\textcolor{red}{-1.1}) & 56.4 (\textcolor{red}{-0.9}) & 34.2 (\textcolor{red}{-1.5}) \\
 ~~w/o AS & 44.3 (\textcolor{red}{-0.6}) & 56.9 (\textcolor{red}{-0.4}) & 35.0 (\textcolor{red}{-0.7}) \\
\bottomrule
\end{tabular}

\caption{
Ablation study of SSV-CoT on commonsense visual reasoning benchmarks.
SR denotes Structured Regions, SS denotes Sequential Selection, and AS denotes Adaptive Stopping.
Numbers in parentheses denote performance degradation compared with the full model.
}

\label{tab:ssv_ablation}
\end{table}

\begin{figure*}[t]
    \centering
    \includegraphics[width=\linewidth]{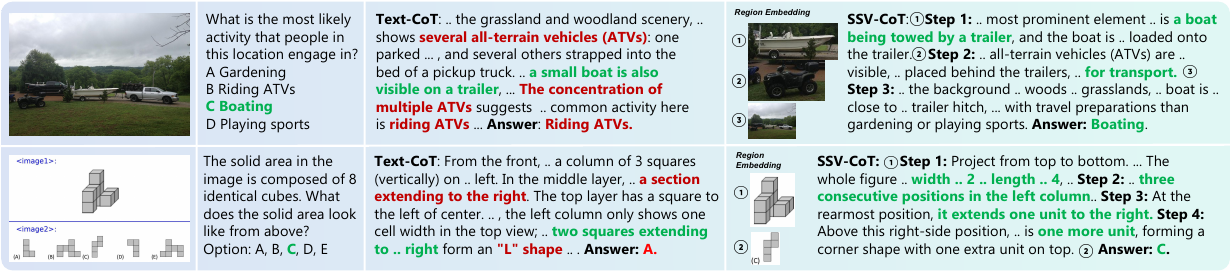}
    \caption{Comparison of Text-CoT and SSV-CoT on Qwen2-VL-7B for visual reasoning. Errors are shown in red. Only selected regions are displayed, and numbers indicate the token injection order.}
    \label{fig:case}
\end{figure*}

\subsection{Ablation Experiment}
We ablate SSV-CoT to assess the contribution of its core components across three commonsense visual reasoning benchmarks (Table~\ref{tab:ssv_ablation}), with the following settings: 
(1) \textit{w/o Structured Regions}: the model does not construct question-conditioned structured regions and directly injects raw visual tokens, resembling patch-level interleaved reasoning; 
(2) \textit{w/o Sequential Selection}: all region embeddings are injected at once, removing the sequential visual access order; 
(3) \textit{w/o Adaptive Stopping}: the model is forced to attend to a fixed number of regions without dynamically deciding when to stop visual querying.

The results show that all components contribute to SSV-CoT. 
Removing structured regions causes the largest performance drop, while disabling sequential selection also consistently degrades performance. 
In contrast, removing adaptive stopping leads to smaller but stable declines. 
Performance drops are more pronounced on M3CoT and LLaVA-W than on ScienceQA, reflecting their stronger reliance on fine-grained visual information.

\begin{table}[t]
\centering
\small
\setlength{\tabcolsep}{6pt}
\begin{tabular*}{\columnwidth}{@{\extracolsep{\fill}}lcc}
\toprule
            & \multicolumn{2}{c}{Order (Acc@1 \%)} \\
\cmidrule(lr){2-3}
Visual repr.& Random & Cognition  \\
\midrule
Patch subset     & 63.1 & 65.4 \\
Saliency regions & 68.0 & 72.2 \\
\bottomrule
\end{tabular*}
\caption{2$\times$2 ablation of visual structure and cognition order on MathVista. Each cell reports Acc@1 (\%). All variants share the same visual token budget and maximum steps $K$.}
\label{tab:struct_order_2x2}
\end{table}

\begin{table}[t]
\centering
\small
\setlength{\tabcolsep}{4pt}
\begin{tabular*}{\columnwidth}{@{\extracolsep{\fill}}lcc}
\toprule
Strategy              & Acc@1 (\%) & Avg \#regions \\
\midrule
Fixed-$K{=}2$ (Seq)   & 68.5       & 2.0 \\
Fixed-$K{=}3$ (Seq)   & 71.0       & 3.0 \\
Fixed-$K{=}4$ (Seq)   & 71.3       & 4.0 \\
Adaptive-$K$ (Ours)   & 72.2       & 2.7 \\
\bottomrule
\end{tabular*}
\caption{Visual budget analysis on MathVista. Fixed-$K$ variants always query exactly $K$ regions. Our adaptive-$K$ policy learns when to stop via a \texttt{<STOP>} action, achieving comparable or better accuracy with fewer region queries on average.}
\label{tab:budget}
\end{table}

\subsection{In-depth Analysis}
\paragraph{Structure and order are mutually reinforcing.}

Table~\ref{tab:struct_order_2x2} jointly ablates visual structure and cognition order on MathVista. Using unstructured patch subsets with a random schedule (\textit{Patch subset, Random}) gives the weakest performance. Introducing a learned access order on top of the same patches (\textit{Patch subset, Cognition}) yields a modest gain (63.1$\rightarrow$65.4), suggesting that sequential, goal-conditioned access helps even with weak visual units. Replacing patches with saliency-based regions under a random order (\textit{Saliency regions, Random}) brings a larger improvement (63.1$\rightarrow$68.0), showing that structured, semantically coherent regions are crucial on their own. Our full model (\textit{Saliency regions, Cognition}) combines both ingredients and achieves the best accuracy (72.2), with gains that exceed either modification in isolation. This pattern indicates that region structure and cognition order are complementary: the policy needs good regions to reason over, and good regions are used most effectively when queried in a learned sequence.

\paragraph{Adaptive visual budget.}
Table~\ref{tab:budget} analyzes how the visual budget interacts with the sequential policy. Increasing the fixed number of region queries from $K{=}2$ to $K{=}3$ substantially improves performance (68.5$\rightarrow$71.0), but further raising the budget to $K{=}4$ yields only marginal gains (71.3) while forcing the model to inspect more regions. In contrast, our adaptive-$K$ variant, equipped with a learned \texttt{<STOP>} action, reaches the highest accuracy (72.2) while using 2.7 regions on average. This shows that the model not only learns \emph{where} and \emph{in which order} to look, but also \emph{how much} visual evidence is needed for a given question, yielding more accurate reasoning with a tighter and automatically controlled visual budget.

\subsection{Case Study}

This Fig. \ref{fig:case} compares the performance of Text-CoT and Structured Visual Reasoning (SSV-CoT) in two typical cases: common-sense visual reasoning (top) and mathematical visual reasoning (bottom). In the common-sense case, Text-CoT focuses on salient but incomplete clues (e.g., multiple ATVs) while ignoring finer contextual details, leading to incorrect conclusions. In contrast, SSV-CoT sequentially focuses on task-relevant areas (boat, trailer, background), progressively integrating evidence to arrive at the correct answer. In the mathematical case, Text-CoT relies on a holistic description of the structure, misinterpreting the spatial layout. However, SSV-CoT explicitly selects and infers key areas (projection, column continuity, extension, and height), enabling accurate spatial reasoning.

\section{Conclusion}

We introduce SSV-CoT, which frames visual cognition in multimodal reasoning as a structured, sequential process instead of static token injection. Trained with chain-of-thought supervision and answer-level rewards, SSV-CoT learns visual selection and stopping without region annotations or external tools. Empirical results across commonsense and math benchmarks demonstrate consistent improvements and highlight the importance of structured regions and adaptive visual access.

\newpage
\section*{Limitations}
SSV-CoT introduces additional computation due to sequential visual access compared with one-shot visual token injection.
Although the vision encoder is executed only once, the model still needs extra region scoring and policy decisions during decoding.
Moreover, its effectiveness depends on the quality of the initially constructed question-conditioned region bank.
If the saliency map misses small but decisive evidence, or if the connected components fail to cover the correct fine-grained region, the sequential policy cannot recover a new region outside the precomputed candidate bank.
This limitation is especially relevant for dense charts, diagrams, and spatial reasoning tasks.
Extending SSV-CoT with multi-scale region proposals, iterative region refinement, or dynamic region generation during reasoning remains future work.

\section*{Ethical Considerations}
SSV-CoT is trained on existing vision-language datasets and does not require additional annotations or external tools.
Like other vision-language models, it may inherit biases from the training data.
Care should be taken when applying the method in real-world or high-stakes scenarios.

\bibliography{acl2020}

@misc{guo2025structuredoutputs,
      title={Structured Outputs Enable General-Purpose LLMs to be Medical Experts}, 
      author={Guangfu Guo and Kai Zhang and others},
      year={2025},
      eprint={2503.03194},
      archivePrefix={arXiv},
      primaryClass={cs.CL},
      url={https://arxiv.org/abs/2503.03194}, 
}

@inproceedings{guo-etal-2025-med,
    title = "{M}ed-{VRA}gent: A Framework for Medical Visual Reasoning-Enhanced Agents",
    author = "Guo, Guangfu  and
      Lu, Xiaoqian  and
      Feng, Yue",
    editor = "Christodoulopoulos, Christos  and
      Chakraborty, Tanmoy  and
      Rose, Carolyn  and
      Peng, Violet",
    booktitle = "Proceedings of the 2025 Conference on Empirical Methods in Natural Language Processing",
    month = nov,
    year = "2025",
    address = "Suzhou, China",
    publisher = "Association for Computational Linguistics",
    url = "https://aclanthology.org/2025.emnlp-main.939/",
    doi = "10.18653/v1/2025.emnlp-main.939",
    pages = "18602--18616",
    ISBN = "979-8-89176-332-6"
}

@book{yarbus1967eye,
  title     = {Eye Movements and Vision},
  author    = {Yarbus, Alfred L.},
  year      = {1967},
  publisher = {Plenum Press},
  address   = {New York}
}

@ARTICLE{itti1998model,
  author={Itti, L. and Koch, C. and Niebur, E.},
  journal={IEEE Transactions on Pattern Analysis and Machine Intelligence}, 
  title={A model of saliency-based visual attention for rapid scene analysis}, 
  year={1998},
  volume={20},
  number={11},
  pages={1254-1259},
  doi={10.1109/34.730558}}

@inproceedings{
schick2023toolformerlanguagemodelsteach,
title={Toolformer: Language Models Can Teach Themselves to Use Tools},
author={Timo Schick and Jane Dwivedi-Yu and others},
booktitle={Thirty-seventh Conference on Neural Information Processing Systems},
year={2023},
url={https://openreview.net/forum?id=Yacmpz84TH}
}

@InProceedings{li2022groundedlanguageimagepretraining,
    author    = {Li, Liunian Harold and Zhang, Pengchuan and others},
    title     = {Grounded Language-Image Pre-Training},
    booktitle = {Proceedings of the IEEE/CVF Conference on Computer Vision and Pattern Recognition (CVPR)},
    month     = {June},
    year      = {2022},
    pages     = {10965-10975}
}

@inproceedings{yang2022unitabunifyingtextbox,
  title     = {UniTAB: Unifying Text and Box Outputs for Grounded Vision-Language Modeling},
  author    = {Yang, Zhengyuan and Gan, Zhe and Wang, Jianfeng and others},
  booktitle = {European Conference on Computer Vision},
  year      = {2022},
  pages     = {521--539}
}

@inproceedings{liu2024groundingdinomarryingdino,
  title     = {Grounding DINO: Marrying DINO with Grounded Pre-Training for Open-Set Object Detection},
  author    = {Liu, Shilong and Zeng, Zhaoyang and Ren, Tianhe and others},
  booktitle = {European Conference on Computer Vision},
  year      = {2024},
  pages     = {38--55}
}

@InProceedings{kirillov2023segment,
    author    = {Kirillov, Alexander and Mintun, Eric and others},
    title     = {Segment Anything},
    booktitle = {Proceedings of the IEEE/CVF International Conference on Computer Vision (ICCV)},
    month     = {October},
    year      = {2023},
    pages     = {4015-4026}
}

@inproceedings{
chen2022regionvitregionaltolocalattentionvision,
title={RegionViT: Regional-to-Local Attention for Vision Transformers},
author={Chun-Fu Chen and Rameswar Panda and Quanfu Fan},
booktitle={International Conference on Learning Representations},
year={2022},
url={https://openreview.net/forum?id=T__V3uLix7V}
}

@article{
zhang2024multimodalchainofthoughtreasoninglanguage,
title={Multimodal Chain-of-Thought Reasoning in Language Models},
author={Zhuosheng Zhang and Aston Zhang and others},
journal={Transactions on Machine Learning Research},
issn={2835-8856},
year={2024},
url={https://openreview.net/forum?id=y1pPWFVfvR},
note={}
}

@inproceedings{
alayrac2022flamingovisuallanguagemodel,
title={Flamingo: a Visual Language Model for Few-Shot Learning},
author={Jean-Baptiste Alayrac and Jeff Donahue and others},
booktitle={Advances in Neural Information Processing Systems},
editor={Alice H. Oh and Alekh Agarwal and Danielle Belgrave and Kyunghyun Cho},
year={2022},
url={https://openreview.net/forum?id=EbMuimAbPbs}
}

@InProceedings{li2023blip2bootstrappinglanguageimagepretraining,
  title = 	 {{BLIP}-2: Bootstrapping Language-Image Pre-training with Frozen Image Encoders and Large Language Models},
  author =       {Li, Junnan and Li, Dongxu and others},
  booktitle = 	 {Proceedings of the 40th International Conference on Machine Learning},
  pages = 	 {19730--19742},
  year = 	 {2023},
  editor = 	 {Krause, Andreas and Brunskill, Emma and others},
  volume = 	 {202},
  series = 	 {Proceedings of Machine Learning Research},
  month = 	 {23--29 Jul},
  publisher =    {PMLR},
  url = 	 {https://proceedings.mlr.press/v202/li23q.html}
}

@inproceedings{
dai2023instructblipgeneralpurposevisionlanguagemodels,
title={Instruct{BLIP}: Towards General-purpose Vision-Language Models with Instruction Tuning},
author={Wenliang Dai and Junnan Li and others},
booktitle={Thirty-seventh Conference on Neural Information Processing Systems},
year={2023},
url={https://openreview.net/forum?id=vvoWPYqZJA}
}

@inproceedings{mnih2014recurrentmodelsvisualattention,
  author       = {Volodymyr Mnih and
                  Nicolas Heess and
                  others},
  editor       = {Zoubin Ghahramani and
                  Max Welling and
                  others},
  title        = {Recurrent Models of Visual Attention},
  booktitle    = {Advances in Neural Information Processing Systems 27: Annual Conference
                  on Neural Information Processing Systems 2014, December 8-13 2014,
                  Montreal, Quebec, Canada},
  pages        = {2204--2212},
  year         = {2014},
  url          = {https://proceedings.neurips.cc/paper/2014/hash/09c6c3783b4a70054da74f2538ed47c6-Abstract.html},
  bibsource    = {dblp computer science bibliography, https://dblp.org}
}

@inproceedings{ba2015multipleobjectrecognitionvisual,
  author       = {Jimmy Ba and
                  Volodymyr Mnih and
                  Koray Kavukcuoglu},
  editor       = {Yoshua Bengio and
                  Yann LeCun},
  title        = {Multiple Object Recognition with Visual Attention},
  booktitle    = {3rd International Conference on Learning Representations, {ICLR} 2015,
                  San Diego, CA, USA, May 7-9, 2015, Conference Track Proceedings},
  year         = {2015},
  url          = {http://arxiv.org/abs/1412.7755},
  bibsource    = {dblp computer science bibliography, https://dblp.org}
}

@inproceedings{
jaegle2022perceiveriogeneralarchitecture,
title={Perceiver {IO}: A General Architecture for Structured Inputs \& Outputs},
author={Andrew Jaegle and Sebastian Borgeaud and others},
booktitle={International Conference on Learning Representations},
year={2022},
url={https://openreview.net/forum?id=fILj7WpI-g}
}

@inproceedings{
wei2023chainofthoughtpromptingelicitsreasoning,
title={Chain of Thought Prompting Elicits Reasoning in Large Language Models},
author={Jason Wei and Xuezhi Wang and others},
booktitle={Advances in Neural Information Processing Systems},
editor={Alice H. Oh and Alekh Agarwal and Danielle Belgrave and Kyunghyun Cho},
year={2022},
url={https://openreview.net/forum?id=_VjQlMeSB_J}
}

@inproceedings{
lu2022learnexplainmultimodalreasoning,
title={Learn to Explain: Multimodal Reasoning via Thought Chains for Science Question Answering},
author={Pan Lu and Swaroop Mishra and others},
booktitle={Advances in Neural Information Processing Systems},
editor={Alice H. Oh and Alekh Agarwal and others},
year={2022},
url={https://openreview.net/forum?id=HjwK-Tc_Bc}
}

@inproceedings{
lu2024mathvistaevaluatingmathematicalreasoning,
title={MathVista: Evaluating Mathematical Reasoning of Foundation Models in Visual Contexts},
author={Pan Lu and Hritik Bansal and others},
booktitle={The Twelfth International Conference on Learning Representations},
year={2024},
url={https://openreview.net/forum?id=KUNzEQMWU7}
}

@inproceedings{
wang2024measuringmultimodalmathematicalreasoning,
title={Measuring Multimodal Mathematical Reasoning with {MATH}-Vision Dataset},
author={Ke Wang and Junting Pan and others},
booktitle={The Thirty-eight Conference on Neural Information Processing Systems Datasets and Benchmarks Track},
year={2024},
url={https://openreview.net/forum?id=QWTCcxMpPA}
}

@misc{chen2023shikraunleashingmultimodalllms,
      title={Shikra: Unleashing Multimodal LLM's Referential Dialogue Magic}, 
      author={Keqin Chen and Zhao Zhang and others},
      year={2023},
      eprint={2306.15195},
      archivePrefix={arXiv},
      primaryClass={cs.CV},
      url={https://arxiv.org/abs/2306.15195}, 
}

@misc{yang2023mmreactpromptingchatgptmultimodal,
      title={MM-REACT: Prompting ChatGPT for Multimodal Reasoning and Action}, 
      author={Zhengyuan Yang and Linjie Li and others},
      year={2023},
      eprint={2303.11381},
      archivePrefix={arXiv},
      primaryClass={cs.CV},
      url={https://arxiv.org/abs/2303.11381}, 
}

@misc{aissi2025vipervisualperceptionexplainable,
      title={VIPER: Visual Perception and Explainable Reasoning for Sequential Decision-Making}, 
      author={Mohamed Salim Aissi and Clemence Grislain and others},
      year={2025},
      eprint={2503.15108},
      archivePrefix={arXiv},
      primaryClass={cs.LG},
      url={https://arxiv.org/abs/2503.15108}, 
}

@InProceedings{kamath2021mdetrmodulateddetection,
    author    = {Kamath, Aishwarya and Singh, Mannat and others},
    title     = {MDETR - Modulated Detection for End-to-End Multi-Modal Understanding},
    booktitle = {Proceedings of the IEEE/CVF International Conference on Computer Vision (ICCV)},
    month     = {October},
    year      = {2021},
    pages     = {1780-1790}
}

@inproceedings{
ravi2024sam2segmentimages,
title={{SAM} 2: Segment Anything in Images and Videos},
author={Nikhila Ravi and Valentin Gabeur and others},
booktitle={The Thirteenth International Conference on Learning Representations},
year={2025},
url={https://openreview.net/forum?id=Ha6RTeWMd0}
}

@inproceedings{
shao2024visualcotadvancingmultimodal,
title={Visual CoT: Advancing Multi-Modal Language Models with a Comprehensive Dataset and Benchmark for Chain-of-Thought Reasoning},
author={Hao Shao and Shengju Qian and others},
booktitle={The Thirty-eight Conference on Neural Information Processing Systems Datasets and Benchmarks Track},
year={2024},
url={https://openreview.net/forum?id=aXeiCbMFFJ}
}

@misc{zhang2024cocotcontrastivechainofthoughtprompting,
      title={CoCoT: Contrastive Chain-of-Thought Prompting for Large Multimodal Models with Multiple Image Inputs}, 
      author={Daoan Zhang and Junming Yang and others},
      year={2024},
      eprint={2401.02582},
      archivePrefix={arXiv},
      primaryClass={cs.CV},
      url={https://arxiv.org/abs/2401.02582}, 
}

@inproceedings{li2025aimcotactiveinformationdrivenmultimodal,
    title = "{AIM}-{C}o{T}: Active Information-driven Multimodal Chain-of-Thought for Vision-Language Reasoning",
    author = "Li, Xiping  and
      Ma, Jianghong",
    editor = "Liakata, Maria  and
      Moreira, Viviane P.  and
      others",
    booktitle = "Proceedings of the 64th Annual Meeting of the {A}ssociation for {C}omputational {L}inguistics (Volume 1: Long Papers)",
    month = jul,
    year = "2026",
    address = "San Diego, California, United States",
    publisher = "Association for Computational Linguistics",
    url = "https://aclanthology.org/2026.acl-long.1227/",
    doi = "10.18653/v1/2026.acl-long.1227",
    pages = "26656--26681",
    ISBN = "979-8-89176-390-6"
}

@inproceedings{
liu2023visualinstructiontuning,
title={Visual Instruction Tuning},
author={Haotian Liu and Chunyuan Li and others},
booktitle={Thirty-seventh Conference on Neural Information Processing Systems},
year={2023},
url={https://openreview.net/forum?id=w0H2xGHlkw}
}

@misc{wu2023visualchatgpttalkingdrawing,
      title={Visual ChatGPT: Talking, Drawing and Editing with Visual Foundation Models}, 
      author={Chenfei Wu and Shengming Yin and others},
      year={2023},
      eprint={2303.04671},
      archivePrefix={arXiv},
      primaryClass={cs.CV},
      url={https://arxiv.org/abs/2303.04671}, 
}

@inproceedings{
lin2022reviveregionalvisualrepresentation,
title={{REVIVE}: Regional Visual Representation Matters in Knowledge-Based Visual Question Answering},
author={Leroy Lin and Yujia Xie and others},
booktitle={Advances in Neural Information Processing Systems},
editor={Alice H. Oh and Alekh Agarwal and others},
year={2022},
url={https://openreview.net/forum?id=wwyiEyK-G5D}
}

@conference{khan2022weaklysupervisedgroundingvqa,
title = {Weakly Supervised Grounding for VQA in Vision-Language Transformers},
author = {Aisha Urooj Khan and Hilde Kuehne and others},
url = {https://www.crcv.ucf.edu/wp-content/uploads/2018/11/1011.pdf
https://www.crcv.ucf.edu/wp-content/uploads/2018/11/1011-supp.pdf
https://github.com/aurooj/WSG-VQA-VLTransformers
https://youtu.be/dekmVb6lq3I},
year = {2022},
date = {2022-10-23},
urldate = {2022-10-23},
booktitle = {European Conference on Computer Vision},
pubstate = {published},
tppubtype = {conference}
}

@InProceedings{esser2021tamingtransformershighresolutionimage,
    author    = {Esser, Patrick and Rombach, Robin and Ommer, Bjorn},
    title     = {Taming Transformers for High-Resolution Image Synthesis},
    booktitle = {Proceedings of the IEEE/CVF Conference on Computer Vision and Pattern Recognition (CVPR)},
    month     = {June},
    year      = {2021},
    pages     = {12873-12883}
}

@inproceedings{
ding2022cogview2fasterbettertexttoimage,
title={CogView2: Faster and Better Text-to-Image Generation via Hierarchical Transformers},
author={Ming Ding and Wendi Zheng and others},
booktitle={Advances in Neural Information Processing Systems},
editor={Alice H. Oh and Alekh Agarwal and others},
year={2022},
url={https://openreview.net/forum?id=GkDbQb6qu_r}
}

@InProceedings{radford2021learningtransferablevisualmodels,
  title = 	 {Learning Transferable Visual Models From Natural Language Supervision},
  author =       {Radford, Alec and Kim, Jong Wook and others},
  booktitle = 	 {Proceedings of the 38th International Conference on Machine Learning},
  pages = 	 {8748--8763},
  year = 	 {2021},
  editor = 	 {Meila, Marina and Zhang, Tong},
  volume = 	 {139},
  series = 	 {Proceedings of Machine Learning Research},
  month = 	 {18--24 Jul},
  publisher =    {PMLR},
  url = 	 {https://proceedings.mlr.press/v139/radford21a.html}
}

@InProceedings{mitra2024compositionalchainofthoughtpromptinglarge,
    author    = {Mitra, Chancharik and Huang, Brandon and others},
    title     = {Compositional Chain-of-Thought Prompting for Large Multimodal Models},
    booktitle = {Proceedings of the IEEE/CVF Conference on Computer Vision and Pattern Recognition (CVPR)},
    month     = {June},
    year      = {2024},
    pages     = {14420-14431}
}

@inproceedings{
zheng2023ddcotdutydistinctchainofthoughtprompting,
title={{DDC}oT: Duty-Distinct Chain-of-Thought Prompting for Multimodal Reasoning in Language Models},
author={Ge Zheng and Bin Yang and others},
booktitle={Thirty-seventh Conference on Neural Information Processing Systems},
year={2023},
url={https://openreview.net/forum?id=ktYjrgOENR}
}

@inproceedings{lei2024scaffoldingcoordinatespromotevisionlanguage,
    title = "Scaffolding Coordinates to Promote Vision-Language Coordination in Large Multi-Modal Models",
    author = "Lei, Xuanyu  and
      Yang, Zonghan  and
      others",
    editor = "Rambow, Owen  and
      Wanner, Leo  and
      others",
    booktitle = "Proceedings of the 31st International Conference on Computational Linguistics",
    month = jan,
    year = "2025",
    address = "Abu Dhabi, UAE",
    publisher = "Association for Computational Linguistics",
    url = "https://aclanthology.org/2025.coling-main.195/",
    pages = "2886--2903"
}

@InProceedings{gao2025interleavedmodalchainofthought,
    author    = {Gao, Jun and Li, Yongqi and others},
    title     = {Interleaved-Modal Chain-of-Thought},
    booktitle = {Proceedings of the IEEE/CVF Conference on Computer Vision and Pattern Recognition (CVPR)},
    month     = {June},
    year      = {2025},
    pages     = {19520-19529}
}

@misc{yao2024minicpmvgpt4vlevelmllm,
      title={MiniCPM-V: A GPT-4V Level MLLM on Your Phone}, 
      author={Yuan Yao and Tianyu Yu and others},
      year={2024},
      eprint={2408.01800},
      archivePrefix={arXiv},
      primaryClass={cs.CV},
      url={https://arxiv.org/abs/2408.01800}, 
}

@inproceedings{
fu2025vita15gpt4olevelrealtime,
title={{VITA}-1.5: Towards {GPT}-4o Level Real-Time Vision and Speech Interaction},
author={Chaoyou Fu and Haojia Lin and others},
booktitle={The Thirty-ninth Annual Conference on Neural Information Processing Systems},
year={2025},
url={https://openreview.net/forum?id=8PUzLga3lU}
}

@InProceedings{xu2025llavacotletvisionlanguage,
    author    = {Xu, Guowei and Jin, Peng and others},
    title     = {LLaVA-CoT: Let Vision Language Models Reason Step-by-Step},
    booktitle = {Proceedings of the IEEE/CVF International Conference on Computer Vision (ICCV)},
    month     = {October},
    year      = {2025},
    pages     = {2087-2098}
}

@misc{wang2024qwen2vlenhancingvisionlanguagemodels,
      title={Qwen2-VL: Enhancing Vision-Language Model's Perception of the World at Any Resolution}, 
      author={Peng Wang and Shuai Bai and others},
      year={2024},
      eprint={2409.12191},
      archivePrefix={arXiv},
      primaryClass={cs.CV},
      url={https://arxiv.org/abs/2409.12191}, 
}

@misc{chen2025expandingperformanceboundariesopensource,
      title={Expanding Performance Boundaries of Open-Source Multimodal Models with Model, Data, and Test-Time Scaling}, 
      author={Zhe Chen and Weiyun Wang and others},
      year={2025},
      eprint={2412.05271},
      archivePrefix={arXiv},
      primaryClass={cs.CV},
      url={https://arxiv.org/abs/2412.05271}, 
}

@inproceedings{sun2025mitigatingvisualforgettingtakealong,
    title = "Mitigating Visual Forgetting via Take-along Visual Conditioning for Multi-modal Long {C}o{T} Reasoning",
    author = "Sun, Hai-Long  and
      Sun, Zhun  and
      others",
    editor = "Che, Wanxiang  and
      Nabende, Joyce  and
      others",
    booktitle = "Proceedings of the 63rd Annual Meeting of the Association for Computational Linguistics (Volume 1: Long Papers)",
    month = jul,
    year = "2025",
    address = "Vienna, Austria",
    publisher = "Association for Computational Linguistics",
    url = "https://aclanthology.org/2025.acl-long.257/",
    doi = "10.18653/v1/2025.acl-long.257",
    pages = "5158--5171",
    ISBN = "979-8-89176-251-0"
}

@misc{liu2024points15buildingvisionlanguagemodel,
      title={POINTS1.5: Building a Vision-Language Model towards Real World Applications}, 
      author={Yuan Liu and Le Tian and others},
      year={2024},
      eprint={2412.08443},
      archivePrefix={arXiv},
      primaryClass={cs.CV},
      url={https://arxiv.org/abs/2412.08443}, 
}

@misc{lu2024ovisstructuralembeddingalignment,
      title={Ovis: Structural Embedding Alignment for Multimodal Large Language Model}, 
      author={Shiyin Lu and Yang Li and others},
      year={2024},
      eprint={2405.20797},
      archivePrefix={arXiv},
      primaryClass={cs.CV},
      url={https://arxiv.org/abs/2405.20797}, 
}

@inproceedings{shi2024mathllavabootstrappingmathematicalreasoning,
    title = "Math-{LL}a{VA}: Bootstrapping Mathematical Reasoning for Multimodal Large Language Models",
    author = "Shi, Wenhao  and
      Hu, Zhiqiang  and
      others",
    editor = "Al-Onaizan, Yaser  and
      Bansal, Mohit  and
      Chen, Yun-Nung",
    booktitle = "Findings of the Association for Computational Linguistics: EMNLP 2024",
    month = nov,
    year = "2024",
    address = "Miami, Florida, USA",
    publisher = "Association for Computational Linguistics",
    url = "https://aclanthology.org/2024.findings-emnlp.268/",
    doi = "10.18653/v1/2024.findings-emnlp.268",
    pages = "4663--4680"
}

@InProceedings{zhang2025r1vl,
    author    = {Zhang, Jingyi and Huang, Jiaxing and others},
    title     = {R1-VL: Learning to Reason with Multimodal Large Language Models via Step-wise Group Relative Policy Optimization},
    booktitle = {Proceedings of the IEEE/CVF International Conference on Computer Vision (ICCV)},
    month     = {October},
    year      = {2025},
    pages     = {1859-1869}
}

@inproceedings{
chen2025mintcot,
title={{MINT}-CoT: Enabling Interleaved Visual Tokens in Mathematical Chain-of-Thought Reasoning},
author={Xinyan Chen and Renrui Zhang and others},
booktitle={The Thirty-ninth Annual Conference on Neural Information Processing Systems},
year={2025},
url={https://openreview.net/forum?id=vMpvtSmtXY}
}

@inproceedings{
yao2024mulberry,
title={Mulberry: Empowering {MLLM} with o1-like Reasoning and Reflection via Collective Monte Carlo Tree Search},
author={Huanjin Yao and Jiaxing Huang and others},
booktitle={The Thirty-ninth Annual Conference on Neural Information Processing Systems},
year={2025},
url={https://openreview.net/forum?id=lwOV2ACEK9}
}

\clearpage
\appendix

\section{ Algorithm } 

\begin{algorithm}[!h]
\small
\caption{Question-aware Discriminative Region Selection with Adaptive Token Compression}
\label{alg:mllm_region_selection}

\textbf{Require:} Image $x$, text query $q$, multimodal model $\mathcal{M}$, number of regions $N$, token budget $n$ \\
\textbf{Ensure:} Region set $\mathcal{R}=\{R_1,\dots,R_N\}$ and region tokens $\{E_k\}_{k=1}^N$

\vspace{0.5em}

$V \leftarrow \text{VisionEncoder}_{\mathcal{M}}(x)$,\;
$t \leftarrow \text{TextEncoder}_{\mathcal{M}}(q)$ \\

$A(i,j) \leftarrow \mathrm{sim}(v_{i,j}, t)$,\;
$\tilde{A} \leftarrow \mathrm{Normalize}(A)$ \\

$M \leftarrow \tilde{A} \ge \mathrm{OtsuThreshold}(\tilde{A})$ \\
$\{R_k\} \leftarrow \mathrm{ConnectedComponents}(M)$ \\

Compute $\rho_k=\frac{1}{|R_k|}\sum_{(i,j)\in R_k}\tilde{A}(i,j)$ \\
Select top $(N\!-\!1)$ regions by $\rho_k$ \\
$R_N \leftarrow \Omega \setminus \bigcup_{k=1}^{N-1} R_k$ \\

\For{each region $R_k$}{
    Collect region tokens $\{v_j\}_{j=1}^{m}$ \\
    \If{$m \le n$}{
        $E_k \leftarrow \{\mathrm{Proj}(v_j)\}_{j=1}^{m}$
    }
    \Else{
        Select top-$n$ tokens by saliency \\
        Cluster selected tokens with $k$-means \\
        $E_k \leftarrow \{\mathrm{Proj}(\bar{v}_i)\}_{i=1}^{n}$
    }
}

\textbf{return} $\{\mathcal{R}, E_k\}_{k=1}^{N}$

\end{algorithm}

Algorithm~\ref{alg:mllm_region_selection} describes the question-aware region selection process. We compute a query-conditioned saliency map over visual patches, apply adaptive thresholding and connected-component analysis to obtain candidate regions, and select the top $N-1$ regions by average saliency, with the remaining area treated as background.

\section{Key Implementation Details}

\paragraph{Region Bank Construction and Step-wise Injection.}
Given an input image $x$ and question $q$, SSV-CoT first constructs a fixed-size candidate region bank
$\mathcal{E}=\{e_1,\dots,e_N\}$ before chain-of-thought decoding.
This stage is performed once per image--question pair: the vision encoder is executed once, and the resulting region embeddings are cached for subsequent reasoning steps.
Importantly, the fixed-size bank does not mean that all region embeddings are injected at the beginning of inference.

During chain-of-thought decoding, whenever a valid reasoning-step boundary is detected, the step-wise visual policy selects one entry from the cached region bank according to the current reasoning state.
The selected region embedding is projected into the LLM hidden space and inserted as additional visual evidence for the next reasoning segment:
\[
\hat{e}_{a_t} = \mathrm{Proj}(e_{a_t}) \in \mathbb{R}^{d_{\mathrm{LM}}}.
\]
If the selected action is \texttt{STOP}, no further visual region is injected and the model proceeds to generate the final answer.
Therefore, what remains fixed during inference is the candidate region bank and its number of slots, while the actually accessed visual evidence is dynamically selected at each reasoning step.
This design neither replaces the original visual tokens nor requires repeatedly running the vision encoder during decoding.

\paragraph{ Region Selection and Hyperparameters.}
We set the default number of regions to $N=5$, consisting of $N\!-\!1$ discriminative regions and one background region. Regions are generated via Otsu-based adaptive thresholding over a similarity heatmap, followed by connected-component analysis. Components with area smaller than $\alpha = 0.01 \times HW$ (where $H,W$ denote the feature map resolution) are discarded. The remaining regions are ranked by their average activation score $\rho_k$, and the top $N\!-\!1$ regions are selected.

\paragraph{ GRPO Reward Design.}
We train the model using GRPO, where the reward is defined as a weighted sum:
\[
r = r_{\text{task}} + \lambda_1 r_{\text{format}} - \lambda_2 r_{\text{length}} - \lambda_3 r_{\text{vision}}.
\]
Here, $r_{\text{task}}$ measures answer correctness, $r_{\text{format}}$ enforces valid output structure, $r_{\text{length}}$ penalizes overly long generations, and $r_{\text{vision}}$ penalizes excessive visual access. We set the weights to
\[
\lambda_1 = 0.2,\quad \lambda_2 = 0.01,\quad \lambda_3 = 0.05.
\]

\paragraph{Training Configuration.}
We use Qwen2-VL-7B as the base multimodal large language model (MLLM) in our experiments. Both projectors are implemented as single-layer linear modules. The training procedure consists of two stages: (1) Supervised Fine-Tuning (SFT), where we train for $3$ epochs with a learning rate of $1\times 10^{-6}$ and a batch size of $64$; and (2) Reinforcement Learning (RL), where we train for $700$ steps with group size $G=4$, a weighting factor $\lambda=0.02$, a learning rate of $1\times 10^{-6}$, and a batch size of $16$. During training, the vision encoder is kept frozen, while all remaining model parameters (including the projector layers) are unfrozen and optimized.

\section{Additional Implementation Clarifications and Analyses}
\label{app:additional_analysis}

\subsection{Similarity Function and Step Boundary}

\paragraph{Cosine similarity for saliency estimation.}
In the main text, the function $\mathrm{sim}(\cdot,\cdot)$ is implemented as cosine similarity. Specifically, given the MLLM-contextualized question representation $q$ and the MLLM-contextualized visual token $\tilde{z}_{ij}$ at spatial position $(i,j)$, the question-conditioned saliency score is computed as
\begin{equation}
S_{ij}
=
\mathrm{cos}(q,\tilde{z}_{ij})
=
\frac{q^\top \tilde{z}_{ij}}
{\|q\|_2 \|\tilde{z}_{ij}\|_2}.
\end{equation}
The resulting similarity scores are then normalized and used for Otsu thresholding, connected-component analysis, and Top-$N$ region selection.

\paragraph{Reasoning-step boundary.}
We use the newline token \texttt{\textbackslash n} as an implicit reasoning-step boundary because the chain-of-thought training data follows a step-wise reasoning template in which each newline typically corresponds to a logical reasoning transition.
To reduce meaningless triggers caused by formatting, visual access is activated only when the generated newline matches the expected step-transition pattern in the prescribed output format.
In addition, the format reward encourages valid reasoning--answer structure, while the visual-access penalty discourages unnecessary region queries.
Nevertheless, this trigger is still a heuristic design. A more robust alternative is to introduce an explicit special step token for visual access control, which we leave for future work.

\subsection{Controlled Same-Training Ablation}
\label{app:same_grpo_ablation}

Public mathematical reasoning baselines may differ in backbone, data scale, supervision signals, and optimization pipelines. Therefore, public benchmark comparisons mainly demonstrate overall competitiveness rather than strict mechanism-level attribution.
To better isolate the contribution of sequential visual access, we conduct a controlled same-training ablation under the same LoRA-GRPO setting.
All variants use the same Qwen2-VL backbone, the same 4K MathV360K SFT data, the same GRPO-LoRA configuration, batch size 16, group size 4, LoRA rank 4, and 300 training steps.
The only difference is whether SSV visual access is introduced and whether the region selection policy is conditioned on the current reasoning state.

\begin{table*}[t]
\centering
\small
\setlength{\tabcolsep}{6pt}
\begin{tabular}{lcc}
\toprule
\textbf{Method} & \textbf{MathVista Acc.} & \textbf{Delta} \\
\midrule
Vanilla Qwen2-VL GRPO-LoRA & 59.5 & -- \\
SSV-GRPO-LoRA w/o step state & 62.0 & +2.5 \\
SSV-GRPO-LoRA & 63.5 & +4.0 \\
\bottomrule
\end{tabular}
\caption{
Controlled same-training ablation on MathVista under the same 4K LoRA-GRPO setting.
The results show that SSV visual access improves over vanilla GRPO, and step-aware sequential selection further improves over non-step-aware visual access.
}
\label{tab:controlled_grpo_ablation}
\end{table*}

As shown in Table~\ref{tab:controlled_grpo_ablation}, SSV-GRPO-LoRA improves over vanilla Qwen2-VL GRPO-LoRA by 4.0 points.
Moreover, removing the step-aware reasoning state reduces the gain from +4.0 to +2.5, indicating that the improvement is not merely due to applying GRPO, but is also related to reasoning-conditioned sequential visual access.

\subsection{Additional Public Mathematical Reasoning Baselines}
\label{app:additional_math_baselines}

We further compare SSV-CoT with recent strong mathematical visual reasoning baselines. These methods differ in training data, supervision type, and optimization recipe, so the comparison should be interpreted as an overall competitiveness comparison rather than a fully controlled attribution experiment.

\begin{table*}[t]
\centering
\small
\setlength{\tabcolsep}{4pt}
\begin{tabular}{lcc}
\toprule
\textbf{Method} & \textbf{Training / Supervision} & \textbf{MathVista Acc.} \\
\midrule
TVC-Qwen2-VL-7B & Large-scale SFT, no RL & 68.1 \\
R1-VL-7B~\citep{zhang2025r1vl} & Mulberry-260K~\citep{yao2024mulberry} + StepGRPO & 69.63 \\
MINT-CoT-7B~\citep{chen2025mintcot} & 54K step-level visual-token annotations + SFT + RL & 73.70 \\
SSV-CoT & 50K+ SFT + GRPO, no region-level annotations & 72.2 \\
\bottomrule
\end{tabular}
\caption{
Additional public mathematical visual reasoning comparison on MathVista.
Unlike methods relying on step-level visual-token annotations, SSV-CoT does not use region-level or visual-token-level annotations.
}
\label{tab:additional_public_math}
\end{table*}

\subsection{Training and Inference Efficiency}
\label{app:efficiency}

SSV-CoT introduces additional region scoring and step-wise visual access, but it does not repeatedly run the vision encoder during decoding.
The image is encoded once before reasoning, and the candidate region bank is cached.
During chain-of-thought generation, the additional computation mainly comes from lightweight similarity scoring over a small region bank and policy selection.

\paragraph{Training overhead.}
We compare Vanilla GRPO-LoRA and SSV-GRPO-LoRA under the same configuration: batch size 16, group size 4, LoRA rank 4, and 300 training updates on NVIDIA H100 GPUs.

\begin{table*}[t]
\centering
\footnotesize
\setlength{\tabcolsep}{3pt}
\begin{tabular}{lccccc}
\toprule
\textbf{Method} & \textbf{Updates} & \textbf{Peak GPU} & \textbf{Mean GPU} & \textbf{Mean Update Wall} & \textbf{Train-loop Wall} \\
 & & \textbf{(GiB)} & \textbf{(GiB)} & \textbf{(s)} & \textbf{(h)} \\
\midrule
Vanilla GRPO-LoRA & 300 & 52.276 & 44.832 & 55.086 & 4.59 \\
SSV-GRPO-LoRA & 300 & 56.481 & 46.165 & 62.086 & 5.18 \\
\bottomrule
\end{tabular}
\caption{
Training efficiency comparison under the same GRPO-LoRA configuration.
SSV-GRPO-LoRA increases mean GPU memory by about 3.0\% and train-loop wall time by about 1.13$\times$.
}
\label{tab:training_efficiency}
\end{table*}

As shown in Table~\ref{tab:training_efficiency}, SSV-GRPO-LoRA introduces a moderate training overhead.
The mean GPU memory increases from 44.832 GiB to 46.165 GiB, and the train-loop wall time increases from 4.59 hours to 5.18 hours.
This overhead mainly comes from region scoring and the step-wise visual access policy, rather than repeated execution of the vision encoder.

\paragraph{Inference latency.}
We also evaluate inference efficiency on 200 MathVista samples using vLLM on a single NVIDIA H100 GPU.

\begin{table*}[t]
\centering
\small
\setlength{\tabcolsep}{5pt}

\begin{tabular*}{\textwidth}{@{\extracolsep{\fill}}lccccc}
\toprule
\multicolumn{6}{c}{\textbf{(a) Inference efficiency on 200 MathVista samples}} \\
\midrule
\textbf{Method} & \textbf{Acc.} & \textbf{Latency / Sample} & \textbf{Extra Latency} & \textbf{Output Tokens} & \textbf{Throughput} \\
 & & \textbf{(ms)} & & & \textbf{(tok/s)} \\
\midrule
Qwen2-VL vLLM & 57.0 & 757.4 & -- & 121.58 & 160.5 \\
SSV-CoT vLLM & 68.0 & 899.0 & +18.7\% & 143.85 & 160.0 \\
\midrule

\multicolumn{6}{c}{\textbf{(b) Adaptive stopping behavior grouped by question complexity}} \\
\midrule
\textbf{Complexity} & \textbf{Samples} & \textbf{Avg. \#Regions} & \textbf{Stop Early $\leq$2} & \textbf{Query More $\geq$3} & \textbf{} \\
\midrule
Simple & 179 & 2.16 & 66.0\% & 34.0\% & \\
Medium & 427 & 2.67 & 43.0\% & 57.0\% & \\
Complex & 394 & 3.05 & 28.0\% & 72.0\% & \\
Overall & 1000 & 2.72 & 41.0\% & 59.0\% & \\
\midrule

\multicolumn{6}{c}{\textbf{(c) Ablation of the global complement region on 200 MathVista samples}} \\
\midrule
\textbf{Method} & \textbf{MathVista Acc.} & \textbf{Avg. \#Regions} & \textbf{Stop Rate} & \textbf{} & \textbf{} \\
\midrule
SSV-CoT w/ global complement & 68.0 & 2.63 & 36.5\% & & \\
SSV-CoT w/o global complement & 66.5 & 2.58 & 42.0\% & & \\
\bottomrule
\end{tabular*}

\caption{
Additional analysis of SSV-CoT.
(a) SSV-CoT improves accuracy with a moderate inference-latency increase while maintaining nearly the same token-level throughput.
(b) The learned \texttt{STOP} action adapts the visual budget according to question complexity.
(c) The global complement region provides modest auxiliary contextual benefit.
}
\label{tab:additional_efficiency_stop_complement}
\end{table*}


Table~\ref{tab:additional_efficiency_stop_complement}(a) shows that SSV-CoT increases end-to-end latency from 757.4 ms to 899.0 ms per sample, while maintaining nearly the same token throughput.
The additional latency is mainly due to region bank construction, region selection, and slightly longer reasoning outputs.

\paragraph{Single-pass visual encoding.}
Let $C_{\mathrm{vis}}$ denote the cost of one vision-encoder forward pass, $C_{\mathrm{dec}}$ denote the language decoding cost, $K$ denote the number of candidate regions, and $T$ denote the number of visual-access decisions.
SSV-CoT performs visual encoding only once and caches the resulting region bank.
Its inference cost can be summarized as
\[
C_{\mathrm{SSV}}
=
C_{\mathrm{vis}}
+
C_{\mathrm{dec}}
+
O(TK d),
\]
where $O(TK d)$ corresponds to similarity scoring between the reasoning-state query and the cached region embeddings.
Since $K$ is small in our implementation, this additional term is lightweight compared with repeatedly processing the image.
In contrast, methods that repeatedly crop, re-encode, or re-inject dense visual inputs may require multiple visual encoding or dense visual-token processing operations during reasoning.
Therefore, the main efficiency advantage of SSV-CoT comes from decoupling one-time region-bank construction from step-wise visual access.

\subsection{Adaptive Stopping Behavior}
\label{app:adaptive_stop}

To verify whether the \texttt{STOP} action learns adaptive stopping rather than converging to a fixed average number of visual queries, we group 1000 evaluation samples by question complexity and report the number of queried regions.

As shown in Table~\ref{tab:additional_efficiency_stop_complement}(b), the average number of queried regions increases from 2.16 for simple questions to 3.05 for complex questions.
Meanwhile, the early-stop ratio decreases from 66.0\% to 28.0\%, while the ratio of querying at least three regions increases from 34.0\% to 72.0\%.
This trend suggests that SSV-CoT dynamically adjusts its visual-access budget according to question complexity.

\subsection{Effect of the Global Complement Region}
\label{app:global_complement}

The global complement region is introduced as an auxiliary robustness mechanism.
It aggregates patches not covered by the selected discriminative regions, providing coarse contextual information from the remaining image area.
It is not intended to fully recover fine-grained evidence missed by the saliency map.

As shown in Table~\ref{tab:additional_efficiency_stop_complement}(c), removing the global complement region decreases accuracy from 68.0\% to 66.5\%.
This indicates that the complement region provides useful auxiliary context.
However, because it is constructed by mean-pooling all unselected patches, it may over-smooth background information, especially when the remaining image contains multiple secondary objects or fine-grained cues.
Therefore, the complement region should be viewed as a coarse robustness component rather than a primary source of fine-grained visual reasoning.

\subsection{Motivation and Limitation of K-means Token Compression}
\label{app:kmeans}

When a candidate region contains no more than $n=48$ visual tokens, SSV-CoT preserves all tokens in that region.
When the number of tokens exceeds the budget, we apply K-means clustering in the embedding space and use the resulting centroids as representative region tokens.
The purpose of this design is to reduce redundant tokens and preserve semantic diversity under a fixed token budget.

Compared with top-saliency selection, K-means avoids over-concentrating on only the strongest local responses and provides broader coverage of the region.
Compared with random sampling, it is more stable and less likely to miss important visual patterns.
However, K-means does not explicitly preserve the original 2D spatial order of visual patches.
Therefore, it is mainly designed for semantic coverage rather than precise geometry preservation.
This limitation may affect tasks requiring highly fine-grained spatial or chart reasoning, and future work may explore spatially constrained clustering or multi-scale region refinement.

\subsection{Failure Case Analysis}
\label{app:failure_case}

Although SSV-CoT mitigates visual forgetting by selecting from a structured region bank, the one-shot construction of this bank can still become a bottleneck.
If the initial saliency-based region proposals miss critical evidence, the sequential policy can only select from the precomputed candidates and cannot recover a new high-resolution region during later reasoning.

For example, in a MathVista document-chart case, the question asks which subject has the highest pulse rate during the baseline period.
The correct answer is subject 1, but SSV-CoT predicts 84.
The selected regions include a coarse page-level region, a narrow strip covering only part of the pulse-rate panel, and an off-target lower-panel crop.
Although the page-level region provides global context, it is too diffuse for fine-grained chart reading.
Meanwhile, the localized regions do not jointly cover the full baseline segment and the corresponding subject-index context.
As a result, the model confuses the subject index with a plotted pulse-rate value.

This failure suggests that region-bank construction is a critical component of SSV-CoT.
When the initially constructed region bank lacks the necessary fine-grained evidence, sequential visual access alone cannot fully recover it.
Future extensions may incorporate multi-scale region proposals, iterative region refinement, or dynamic region generation during reasoning.

\begin{figure*}[t]
\centering
\small

\begin{tcolorbox}[
    enhanced,
    width=\textwidth,
    colback=white,
    colframe=black,
    boxrule=0.6pt,
    arc=2mm,
    left=5pt,
    right=5pt,
    top=5pt,
    bottom=5pt
]

{\large\textbf{Failure Case: Region-bank construction misses fine-grained chart evidence}}
\vspace{0.6em}

\noindent
\begin{minipage}[t]{0.54\textwidth}
    \vspace{0pt}
    \centering
    \includegraphics[
        width=\linewidth,
        height=0.50\textheight,
        keepaspectratio
    ]{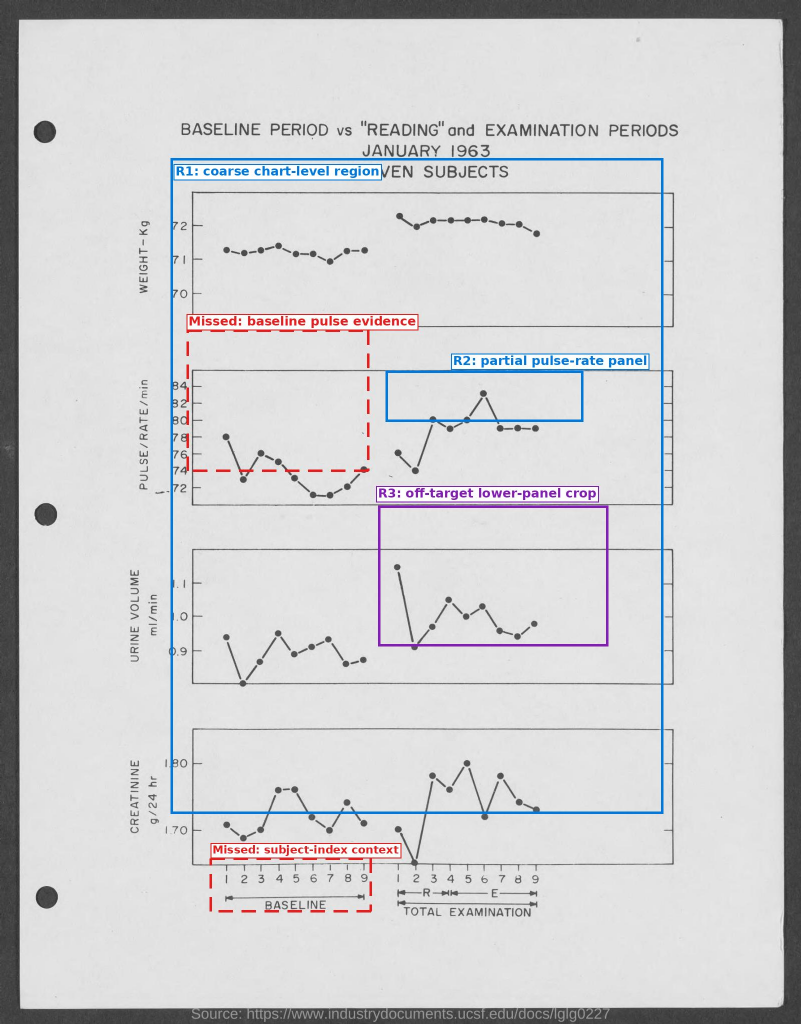}

    \vspace{0.25em}
    {\footnotesize
    Solid boxes: selected regions\\
    Dashed red boxes: missed fine-grained evidence
    }
\end{minipage}
\hfill
\begin{minipage}[t]{0.43\textwidth}
    \vspace{0pt}

\begin{tcolorbox}[
    colback=gray!8,
    colframe=gray!60,
    boxrule=0.4pt,
    arc=1mm,
    left=4pt,
    right=4pt,
    top=4pt,
    bottom=4pt,
    before skip=0pt,
    after skip=0pt
]
\textbf{Input Question}

\vspace{0.25em}
Which subject has the highest pulse rate during the baseline period?

\vspace{0.45em}
\textbf{Ground truth:} Subject 1
\end{tcolorbox}

\vspace{2em}

\begin{tcolorbox}[
    colback=blue!5,
    colframe=blue!45!black,
    boxrule=0.4pt,
    arc=1mm,
    left=4pt,
    right=4pt,
    top=4pt,
    bottom=4pt,
    before skip=0pt,
    after skip=0pt
]
\textbf{Selected Region Bank}

\vspace{0.25em}
\textbf{Region 1:} Coarse page-level region, providing global context but too diffuse for precise chart reading.

\vspace{0.3em}
\textbf{Region 2:} Partial pulse-rate panel, covering only part of the baseline segment and missing subject-index context.

\vspace{0.3em}
\textbf{Region 3:} Off-target lower-panel crop, containing weakly related visual evidence.
\end{tcolorbox}

\vspace{2em}

\begin{tcolorbox}[
    colback=red!5,
    colframe=red!60!black,
    boxrule=0.4pt,
    arc=1mm,
    left=4pt,
    right=4pt,
    top=4pt,
    bottom=4pt,
    before skip=0pt,
    after skip=0pt
]
\textbf{Reasoning Outcome}

\vspace{0.25em}
\textbf{Model prediction:} 84

\vspace{0.45em}
\textbf{Error:} The model confuses a plotted pulse-rate value with the subject index.

\vspace{0.45em}
\textbf{Cause:} The policy can only select from the precomputed region bank; missing fine-grained evidence cannot be recovered during later reasoning.
\end{tcolorbox}

\end{minipage}

\vspace{0.65em}

\begin{tcolorbox}[
    colback=yellow!8,
    colframe=yellow!50!black,
    boxrule=0.4pt,
    arc=1mm,
    left=5pt,
    right=5pt,
    top=4pt,
    bottom=4pt
]
\textbf{Takeaway.}
This case shows that one-shot region-bank construction can become a bottleneck for dense chart reasoning.
When critical fine-grained evidence is not included in the initial candidate regions, the sequential visual policy cannot generate a new high-resolution region during later reasoning.
This motivates future extensions such as multi-scale region proposals, iterative region refinement, or dynamic region generation during CoT reasoning.
\end{tcolorbox}

\end{tcolorbox}

\caption{
A representative failure case caused by imperfect region-bank construction.
Although SSV-CoT performs sequential visual access, the policy can only choose from the precomputed candidate region bank.
If the bank misses critical fine-grained chart evidence, the model may still produce an incorrect answer.
}
\label{fig:failure_case}
\end{figure*}

\end{document}